\documentclass[conference]{IEEEtran}
\IEEEoverridecommandlockouts
\usepackage{cite}
\usepackage{amsmath,amssymb,amsfonts}
\usepackage{graphicx}
\usepackage{hyperref}
\usepackage{textcomp}
\usepackage{xcolor}
\usepackage{algorithm}
\usepackage{adjustbox}
\usepackage{multirow}
\usepackage{array}
\usepackage{nicefrac}
\usepackage{algpseudocode}
\algtext*{EndWhile}% Remove "end while" text
\algtext*{EndIf}% Remove "end if" text
\algtext*{EndFor}% Remove "end for" text
\usepackage{comment}
\def\BibTeX{{\rm B\kern-.05em{\sc i\kern-.025em b}\kern-.08em
    T\kern-.1667em\lower.7ex\hbox{E}\kern-.125emX}}

\definecolor{darkspringgreen}{rgb}{0.09, 0.45, 0.27}

\newcommand{\StatexIndent}[1][3]{%
  \setlength\@tempdima{\algorithmicindent}%
  \Statex\hskip\dimexpr#1\@tempdima\relax}

\newcommand{\rev}[1]{{#1}}

\newcommand{\revFloatRemove}[1]{}
\newcommand{\revEqRemove}[1]{}

\usepackage{tikz}
\usepackage{textcomp}
\usepackage[doipre={DOI:~}]{uri}
\usepackage{lipsum}

\newcommand\copyrighttext{%
  \footnotesize \textcopyright 2022 IEEE. Personal use of this material is permitted.  Permission from IEEE must be obtained for all other uses, in any current or future media, including reprinting/republishing this material for advertising or promotional purposes, creating new collective works, for resale or redistribution to servers or lists, or reuse of any copyrighted component of this work in other works.
 
  Accepted at 2022 IEEE 13th International Green and Sustainable Computing Conference (IGSC).
  ~\doi{10.1109/IGSC55832.2022.9969373} }
\newcommand{\copyrightnotice}{%
\begin{tikzpicture}[remember picture,overlay,scale=1.00, every node/.style={scale=1.00}]
\node[anchor=south,yshift=10pt] at (current page.south) {\fbox{\parbox{\dimexpr\textwidth-\fboxsep-\fboxrule\relax}{\copyrighttext}}};
\end{tikzpicture}%
}

\begin{document}
\bstctlcite{IEEEexample:BSTcontrol}

\title{Channel-wise Mixed-precision Assignment for DNN Inference on Constrained Edge Nodes
\vspace{-.4cm}
}
\author{\IEEEauthorblockN{Matteo Risso\IEEEauthorrefmark{2}, Alessio Burrello\IEEEauthorrefmark{1}, Luca Benini\IEEEauthorrefmark{1}, Enrico Macii\IEEEauthorrefmark{3}, Massimo Poncino\IEEEauthorrefmark{2}, Daniele Jahier Pagliari\IEEEauthorrefmark{2}}
\IEEEauthorblockA{\IEEEauthorrefmark{1}Department of Electrical, Electronic and Information Engineering, University of Bologna, 40136 Bologna, Italy\\
\IEEEauthorrefmark{2}Department of Control and Computer Engineering, Politecnico di Torino, Turin, Italy\\
\IEEEauthorrefmark{3}Inter-university Department of Regional and Urban Studies and Planning, Politecnico di Torino, Turin, Italy\\
%\IEEEauthorrefmark{4}Department of Information Technology and Electrical Engineering at the ETH Zurich, 8092 Zurich, Switzerland\\
}
\IEEEauthorblockA{Corresponding Email: matteo.risso@polito.it}
\vspace{-1.1cm}
}

\maketitle

\copyrightnotice

\begin{abstract}
Quantization is widely employed in both cloud and edge systems to reduce the memory occupation, latency, and energy consumption of deep neural networks. In particular, \textit{mixed-precision} quantization, i.e., the use of different bit-widths for different portions of the network, has been shown to provide excellent efficiency gains with limited accuracy drops, especially with optimized bit-width assignments determined by automated Neural Architecture Search (NAS) tools.
State-of-the-art mixed-precision works \textit{layer-wise}, i.e., it uses different bit-widths for the weights and activations tensors of each network layer.
In this work, we widen the search space, proposing a novel NAS that selects the bit-width of each weight tensor \textit{channel} independently. This gives the tool the additional flexibility of assigning a higher precision only to the weights associated with the most informative features.
Testing on the MLPerf Tiny benchmark suite, we obtain a rich collection of Pareto-optimal models in the accuracy vs model size and accuracy vs energy spaces.
When deployed on the MPIC RISC-V edge processor, our networks reduce the memory and energy for inference by up to $63\%$ and $27\%$ respectively compared to a layer-wise approach, for the same accuracy.

\end{abstract}

\begin{IEEEkeywords}
Deep Learning, NAS, Quantization, TinyML
\end{IEEEkeywords}

\section{Introduction}\label{sec:intro}
Deep Learning (DL) models can enhance the ``smartness' of many edge systems, such as drones~\cite{palossi2022}, wearables~\cite{burrello2021q}, and smart assistants~\cite{zhang2017hello}.
However, the traditional paradigm based on offloading Deep Neural Networks (DNNs) execution to the cloud has several limitations in terms of latency, energy efficiency, and privacy, as it relies on continuous data exchange over the network, often through an unpredictable, unreliable and energy hungry wireless channel~\cite{Daghero2021Energy}. 
Furthermore, the huge carbon footprint of data center \rev{processing} makes the sustainability of this approach questionable~\cite{sustcomp}.

These limitations have spurred an increasing academic and industrial interest for Tiny Machine Learning (\textit{TinyML}), i.e., the study of efficiency-driven optimizations of ML and DL models aimed at making their energy and memory requirements compatible with the tight constraints of edge devices.
\textit{Quantization} is a key step of most TinyML pipelines, which takes advantage of the inherent resilience of DNNs to replace the floating point data representations typically used during training with integers on a limited number of bits~\cite{Jacob2018}.
This is extremely beneficial for edge deployment, as it reduces the memory footprint of the model and avoids the use of slow, power-hungry floating point operations~\cite{Jacob2018}.

Conventional quantization strategies use the same precision for the entire DNN (so-called \textit{fixed-precision})~\cite{choi2018pact, choukroun2019low}, but several recent works have shown that a \textit{mixed-precision} scheme, using different bit-width for different network portions, can further reduce the DNN complexity without accuracy drops~\cite{burrello2021q,cai2020rethinking}. 
However, the additional flexibility of mixed-precision comes with the new challenging problem of optimizing the \textit{bit-width assignment} to different parts of the network. Partially to limit the scope of such optimization, state-of-the-art mixed-precision solutions currently assign bit-widths \textit{layer-wise}. That is, the data representation is optimized independently for each network layer, possibly differently for weights and activations~\cite{cai2020rethinking}.
Even in this setting, the search space is huge; for example, considering a small MobileNet V1~\cite{howard2017mobilenets} and 2/4/8-bit as available bit-widths, the number of possible solutions is $10^{26}$.
Accordingly, a proper optimization cannot be performed by hand, but requires automated Neural Architecture Search (NAS) tools. Recent works such as \cite{wang2019haq, cai2020rethinking} have therefore proposed different optimization methods to find precision assignments offering good trade-offs between the task accuracy and the total memory occupation of the network, or the latency on the final deployment target.

In this work, we assess for the first time the potential of an even finer-grain \textit{channel-wise mixed-precision assignment}. That is, we explore the space of all possible bit-width assignments to each channel of each weight tensor in a Convolutional Neural Network (CNN), while maintaining layer-wise quantization \rev{granularity} for activations. We achieve this goal thanks to a light-weight gradient-based search method, which belongs to the family of \textit{Differentiable} NAS (DNAS). Our channel-wise scheme is fully compatible with existing hardware and software libraries, since the resulting multi-weights-precision convolutions can be easily reconverted into multiple smaller convolutions working in parallel.
With experiments on the four benchmarks of the MLPerf Tiny suite~\cite{banbury2021mlperf} we show that our tool is able to find a rich front of Pareto-optimal solutions.
When deployed on the MPIC~\cite{mpic} RISC-V edge processor, the networks found by our DNAS are able to reduce memory footprint/energy up to \textbf{63\%}/\textbf{27\%} at iso-accuracy with respect to a per-layer mixed-precision search. 
Our code is open-sourced at: \texttt{https://github.com/eml-eda/multi-prec-nas}.

\section{Background and Related Works}
\textbf{Quantization.} 
Quantization is used ubiquitously in TinyML to improve DNN compression and energy-efficiency~\cite{Jacob2018}.
Quantization-aware training (QAT)~\cite{choi2018pact} is a de-facto standard technique that simulates the effect of quantization during training, through so-called \textit{fake-quantization} operations, in order to improve the final accuracy of the quantized model.
Post-training quantization~\cite{capotondi2020cmix} is a viable alternative when re-training is not possible.

\looseness=-1
This work targets the popular \textit{affine quantization} scheme~\cite{Jacob2018}, that maps float tensors $\mathbf{t}$ to $n$-bit integers as:
\begin{equation}
    \mathbf{t}_n = \underset{0:2^{n}-1}{\mathrm{clamp}}\left( \mathrm{round}\left( \frac{\mathbf{t} - \alpha_\mathbf{t}}{\varepsilon_\mathbf{t}} \right)\right) \label{eq:quantz} 
\end{equation}
where $\mathrm{clamp}$ simply restricts the values to $[0:2^{n}-1]$, $[\alpha_\mathbf{t},\beta_\mathbf{t})$ is the range of values that can be mapped without saturation, and $\varepsilon_\mathbf{t} = (\beta_\mathbf{t}-\alpha_\mathbf{t}) / (2^{n}-1)$ is the quantization step.
Notable works that explore this quantization scheme are LQ-Net \cite{zhang2018lq} and PACT\cite{choi2018pact}, which learn the optimal step and 
value range
during training.
Note that our mixed-precision NAS can be be easily extended to non-linear quantization schemes~\cite{han2016deep}, but this paper focuses on affine quantization due to its hardware \rev{friendliness}, which results in a lightweight implementation on most deployment targets~\cite{Jacob2018}.
\textbf{Neural Architecture Search.}
NAS tools are usually designed to explore the space of possible network \textit{topologies} (e.g., selecting among alternative layers or tuning their geometric hyper-parameters). In that, they constitute a more effective alternative to the manual rules of thumb derived from experience that drove the design of early efficient DNNs~\cite{howard2017mobilenets}.

In particular, NAS tools oriented at TinyML co-optimize the network performance (e.g., classification accuracy) and some cost metric (e.g., memory occupation, latency or energy consumption).
Seminal approaches leverage an iterative procedure consisting of: i) sampling one ore more architectures, ii) training them to estimate task accuracy and iii) profiling their cost either through direct deployment or through some proxy model~\cite{nas_evol,mnasnet_2019}. Sampling is driven by black-box optimizers such as Evolutionary Algorithms (EA)~\cite{nas_evol} or Reinforcement Learning (RL)~\cite{mnasnet_2019}.
While powerful, these algorithms require thousands of iterations (hence GPU hours) per search, impairing their sustainability and their.

To cope with these limitations, Differentiable NAS (DNAS) techniques use gradient descent to simultaneously train the network and optimize its architecture, relaxing the topology optimization problem to make it differentiable~\cite{liu2018darts}.
Accuracy/cost co-optimization is achieved adding to the standard training loss a regularization term $\mathcal{L_R}$, modeling the target cost metric (size, energy, etc) in a differentiable way. Thus, the loss function becomes: 
\begin{equation}\label{eq:dnas_loss}
\mathcal{L}(W; \theta) = \mathcal{L_T}(W; \theta) + \lambda \mathcal{L_R}(\theta)
\end{equation}
where $\mathcal{L_T}$ is the standard task loss, $W$ is the set of network weights, and $\theta$ is the set of NAS parameters that determine the network topology. The scalar regularization strength $\lambda$ can be tuned to guide the research towards more accurate or more lightweight networks. The encoding of alternative topologies is obtained from the values $\theta$ either: i) using them to perform a soft-selection among alternative versions of each layer, typically through a softmax (so-called \textit{super-net} approaches)~\cite{liu2018darts,cai2018proxylessnas}, or ii) using them as masks to prune away parts of a large seed model, e.g., some channels in a Conv. layer (so-called \textit{mask-based} approaches)~\cite{wan2020fbnetv2,risso2022PIT}.

\textbf{Mixed-Precision Assignment.}
\looseness=-1
Recently, NAS techniques have also been employed to address the bit-width assignment problem in mixed-precision DNNs. Similarly to topology-NAS, the goal is to simultaneously maximize accuracy and minimize inference costs.
Some methods, such as RELEQ~\cite{elthakeb2019releq} and HAQ~\cite{wang2019haq} use RL agents. While the black-box optimizer allows them to guide the precision search with actual latency or energy measurements from the target hardware, these methods suffer from the same efficiency downsides described previously for other RL-based NAS.
The authors of EdMIPS~\cite{cai2020rethinking}, instead, proposed the first DNAS for mixed-precision assignment. To learn the optimal precision, EdMIPS \textit{simulates} the effect of quantization at different bit-widths for the input activations and weights of each layer, similarly to fake-quantization in QAT~\cite{Jacob2018}. The fake-quantized tensors are combined through ``softmax-ed'' NAS parameters, optimized during training. Using the formulation of (\ref{eq:dnas_loss}), gradient-descent is encouraged to increase the parameters associated with lower bit-widths, if doing so does not harm accuracy. Since our method is inspired by~\cite{cai2020rethinking} more details are provided in Sec.~\ref{sec:method}.

\section{Proposed Method}\label{sec:method}
Mixed-Precision NAS methods like HAQ~\cite{wang2019haq} and EdMIPS~\cite{cai2020rethinking} assign an independent precision to the weights and activations of each network \textit{layer}.
Their rationale is that different layers exhibit different degrees of redundancy and feature extraction power, thus using the same precision for the entire network would be sub-optimal.

In this work we push this idea forward, proposing a lightweight DNAS method able to learn an independent precision assignment for each weight tensor \textit{channel} in convolutional (Conv) or fully-connected (FC) layers, i.e., the two layer types that mostly influence the inference complexity of CNNs. 
In other words, we assign an independent precision to the weights of \textit{each filter} in Conv layers\footnote{\rev{By filter, we mean a slice of weights that processes all channels of an input activation patch, and produces a single output channel.}}, and to the weights associated with \textit{each output neuron} in FC layers. 
We maintain the activation bit-width assignment layer-wise, for implementation reasons clarified below.
In the following, we discuss the details of our method for Conv layers only, since the extension to FC is straight-forward.

Our channel-based precision assignment explores a larger and finer-grain solution space. For instance, considering the same MobileNetV1 mentioned in Sec~\ref{sec:intro} and a width multiplier of $0.25$, the number of solutions grows from $10^{26}$ in a layer-wise approach to $10^{74}$ in ours.
However, this allows our method to exploit the \textit{different relative importance of extracted features} within a single layer to further optimize the model.
\subsection{Precision Assignment Optimization Method} \label{sec:chprec_dnas}

\begin{figure}[t]
  \centering
  \includegraphics[width=\columnwidth]{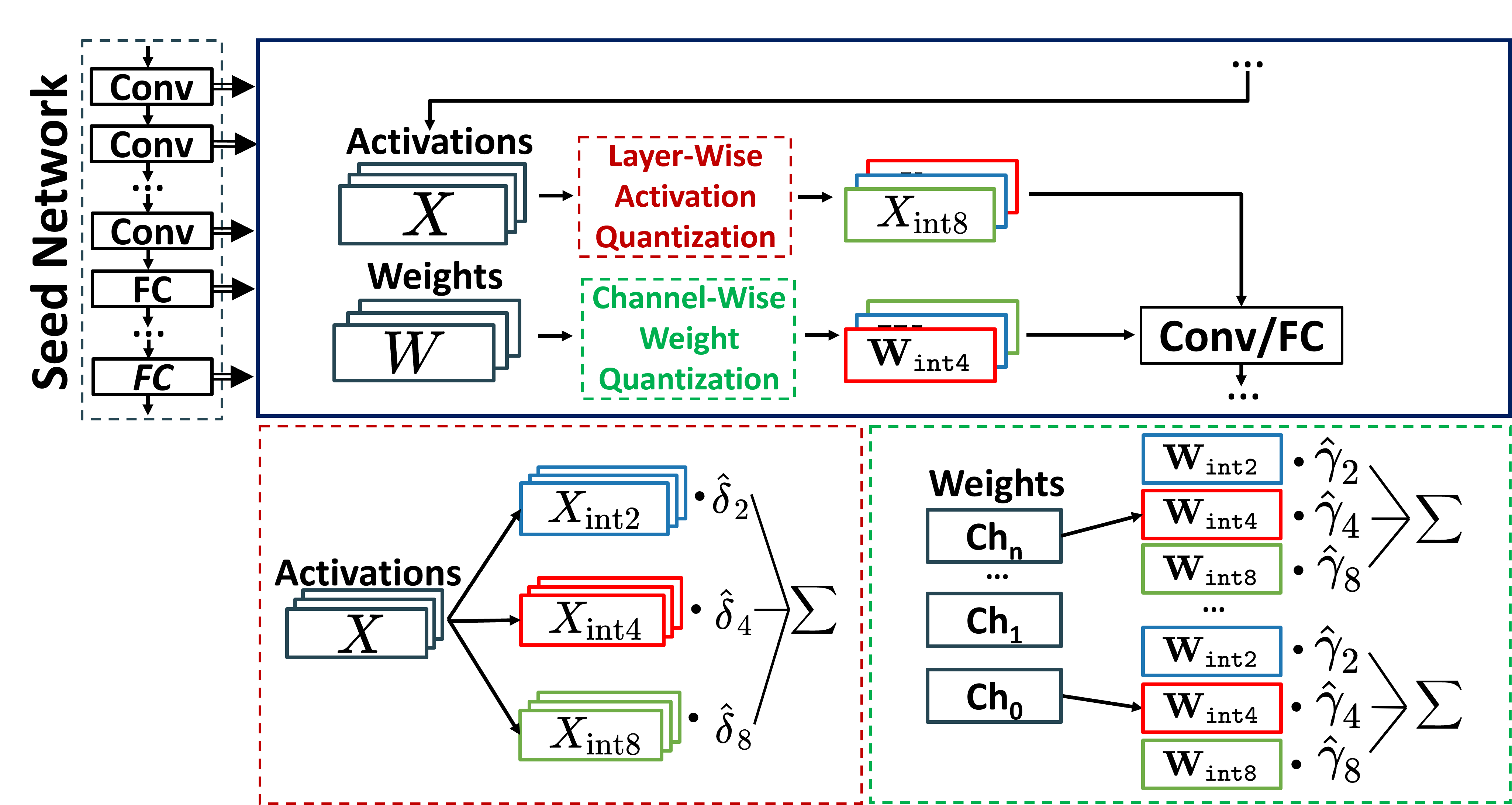}
  \vspace{-0.4cm}
  \caption{Overview of the proposed approach.}
  \vspace{-0.4cm}
  \label{fig:flow}
\end{figure}

Fig.~\ref{fig:flow} summarizes the flow of our approach. 
For each optimized layer of the target DNN, we first fake-quantize the activation tensor $X$ at all supported bit-widths $p_x \in P_x$, e.g., $P_x = \{2, 4, 8\}$ bit. We then combine the fake-quantized tensors through a vector of NAS parameters $\delta^{(n)} \in \mathbb{R}^{|P_x|}$, where the $n$ superscript refers to the $n$-th layer, in a way similar to~\cite{cai2020rethinking}. 
Specifically, we first compute $\hat{\delta}^{(n)} = \mathrm{SM}(\delta^{(n)}; \tau)$ where $\mathrm{SM}(x; \tau)$ is the softmax with temperature:
\begin{equation}\label{eq:softmax}
    SM(x; \tau) = \frac{e^{x_i} / \tau}{\sum_i e^{x_i} / \tau},\ \forall i
\end{equation}
so that the elements of $\hat{\delta}^{(n)}$ sum to 1. As detailed in Sec.~\ref{sec:train_proc}, the temperature $\tau$ is progressively annealed during training, driving the softmax to increasingly resemble a non-differentiable \textit{argmax}, which is the function used to select the final precision at the end of the optimization.

Then, the \textit{effective} activation tensor is obtained as:
\begin{equation}\label{eq:ed_mips}
    \hat{X}^{(n)} = \sum_{p_x \in P_X} \hat{\delta}^{(n)}_{p_x} \cdot X_{p_x}
\end{equation}
where $X_{p_x}$ is the $p_x$-bit fake-quantized version of the original float tensor $X$, and we loosely use $p_x$ to indicate both a precision and an index in the $\hat{\delta}$ array. Similarly to~\cite{cai2020rethinking}, the rationale of (\ref{eq:ed_mips}) is that the larger $\hat{\delta}_{p_x}^{(n)}$, the more $\hat{X}^{(n)}$ becomes similar to the result of a $p_x$-bit quantization.

To explore the channel-wise assignments of bit-widths $p_w \in P_W$ for weights, which is the main novelty of our work, we further associate each layer with a 2D matrix $\gamma^{(n)} \in \mathbb{R}^{C^{(n)}_{out} \times |P_W|}$, where $C^{(n)}_{out}$ is the number of output channels/filters in the layer.
The $i$-th row of the matrix $\gamma^{(n)}_i$ contains the NAS parameters that will determine the bit-width assignment for the $i$-th channel.
Accordingly, each row is applied an independent softmax to obtain $\hat{\gamma}^{(n)}_i = SM(\gamma^{(n)}_i; \tau)$.
Next, the $i$-th effective weight tensor slice $\hat{W}^{(n)}_i$, i.e., the $i$-th effective filter, is obtained similarly to (\ref{eq:ed_mips}), as:
\begin{equation}
\hat{W}^{(n)}_i = \sum_{p_w \in P_W} \hat{\gamma}^{(n)}_{i,p_w} \cdot W^{(n)}_{i,p_w}
\end{equation}
Lastly, the stack of these slices along the $C_{out}$ dimension is used, together with $\hat{X}^{(n)}$ to produce the layer output:
\begin{equation}
Y^{(n)} = \mathrm{Conv}(\hat{X}^{(n)}, \underset{i}{\mathrm{stack}}(\hat{W}^{(n)}_i))
\end{equation}

\looseness=-1
Our method uses \textit{weight sharing}, that is, all fake-quantized weights and activations are obtained from a \textit{single} float tensor, and are generated just once per-layer on the fly, i.e., we have $|P_W|$ and $|P_X|$ temporary copies of $X$ and $W$ during forward DNN passes. Thus, the memory overhead of our method
during training
is almost identical to~\cite{cai2020rethinking}, except for the new $\gamma^{(n)}$ matrix, whose impact is negligible. A key difference between our method and~\cite{cai2020rethinking} is instead in the quantization scheme, i.e., the function mapping $X \rightarrow X_{p_x}$ and $W \rightarrow W_{p_w}$. Namely, we replace the original Gaussian quantizer used in~\cite{cai2020rethinking} with the PaCT method described in~\cite{choi2018pact}. The main reason is that PaCT layers are fully compatible with our deployment target~\cite{mpic}. Further, we found that it also yields superior results.

We optimize the precision assignment while training the DNN weights by minimizing the loss formulation of (\ref{eq:dnas_loss}), where $\theta = \{\delta^{(1)},...,\delta^{(n)}, \gamma^{(1)},...,\gamma^{(n)}\}$ is the set of NAS trainable parameters for all layers.

We use two different expressions for the complexity regularizer $\mathcal{L_R}$ depending on the optimization objective.
Specifically, when the goal is to minimize the \textit{model size}, i.e., the total amount of non-volatile memory required to store the network, the NAS should be guided to select smaller bit-widths for weights tensors channels where possible, whereas activations bit-widths have no impact.
A simple regularizer that achieves this objective for a single Conv layer is:
\begin{equation} \label{eq:reg_loss_size}
    \mathcal{L_R}^{(n)} = C^{(n)}_{in}K^{(n)}_{x}K^{(n)}_{y} \sum_{i=1}^{C^{(n)}_{out}} \sum_{p_{w} \in P_W} \hat{\gamma}^{(n)}_{i,p_{w}} \cdot p_{w}
\end{equation}
\looseness=-1
where $C^{(n)}_{in}$ $K^{(n)}_{x}$ and $K^{(n)}_{y}$ are number of input channels and the horizontal and vertical convolution kernel sizes. In practice, this regularizer computes the \textit{effective number of weight bits} in the layer, multiplying each softmax coefficient $\hat{\gamma}^{(n)}_{i,p_{w}}$ times the corresponding bit-width $p_{w}$. Thus, for instance, it assigns a double cost to the coefficient associated with 4bit precision with respect to the 2bit one. Note that when optimizing the model size, we disable the search over activation bit-widths and fix all $\hat{X}^{(n)}$ tensors to 8bit.

We also consider the optimization of the \textit{energy consumption} of the network, which depends both on weights and activations precision.
The corresponding regularizer is:
\begin{equation} \label{eq:reg_loss_energy}
    \mathcal{L_R}^{(n)} = \Omega^{(n)}\cdot \sum_{p_{x} \in P_X} \hat{\delta}^{(n)}_{p_{x}} \sum_{i=1}^{C^{(n)}_{out}} \sum_{p_{w} \in P_W} \hat{\gamma}^{(n)}_{i, p_{w}} \mathrm{C}(p_{x}, p_{w})
\end{equation}
\looseness=-1
where $\Omega^{(n)}$ is the total number of operations required to produce the $n$-th layer output, which is independent from the precision assignment and can be computed from well-known formulas
for Conv or FC layers.
The rest of (\ref{eq:reg_loss_energy}) computes the \textit{expected average energy per operation} of the layer.
Specifically, $\mathrm{C}(p_{x}, p_{w})$ is a Look-Up Table (LUT) returning an estimate of the energy/OP of a $p_x$-bit $\times p_w$-bit convolution. The LUT is populated profiling the target hardware, and is necessary because for most hardware platforms, the energy cost of arithmetic operations at sub-byte precision is not linearly proportional to the bit-width. The regularizer simply weighs each combination of NAS parameters for activations and weights with the cost of the corresponding mixed-precision operation.

The overall regularization loss for the NAS is obtained summing either (\ref{eq:reg_loss_size}) or (\ref{eq:reg_loss_energy}) over all layers. 

\subsection{Training Procedure} \label{sec:train_proc}
\begin{algorithm}[t]
\begin{algorithmic}[1]
\footnotesize
\caption{\label{alg:nas_search}}
    \For{$i \gets 1, \dots, \rm Epochs_{wu}$} {\color{darkspringgreen}\# warmup loop}
        \State Update $W$ based on $\nabla_{W} \mathcal{L_T}(W_{p_{max}})$
    \EndFor
    \While{not converged} {\color{darkspringgreen}\# search loop}
        \If{$\rm \#samples < 20\% \text{ current epoch}$}
            \State Update $\theta$ based on $\nabla_{\theta} (\mathcal{L_T}(W;\theta) + \lambda \mathcal{L_R}(\theta))$
        \Else
            \State Update $W$ based on $\nabla_{W} (\mathcal{L_T}(W; \theta))$
        \EndIf
        \State Anneal temperature $\tau$
    \EndWhile
    \For{$i \gets 1, \dots, \rm Epochs_{ft}$} {\color{darkspringgreen}\# fine-tuning loop}
        \State Freeze $\theta$ 
        \State Update $W$ based on $\nabla_{W} \mathcal{L_T}(W)$
    \EndFor
\end{algorithmic}
\end{algorithm}
Alg.~\ref{alg:nas_search} shows the training scheme of our method, which is made of three distinct phases.
In the initial \textit{warmup} phase, the network is trained normally with QAT at the maximum supported precision $p_{max}$, while keeping NAS parameters frozen. Thus, only the task-specific loss function $\mathcal{L_T}$ and the normal $W$ weights are optimized. 
In all our experiments, we consider $p_{max} = 8$bit.
Warmup needs to be performed only once, reusing the result for multiple searches.

\looseness=-1
The second phase represents the core of the optimization.
In it, both weights $W$ and NAS parameters $\theta$ are trained to minimize (\ref{eq:dnas_loss}).
Following the scheme proposed in~\cite{wan2020fbnetv2} we train NAS parameters and normal weights in an alternated fashion within each epoch.
First, only the NAS parameters are trained on a random 20\% split of the training samples.
Then, only the network weights are trained on the remaining 80\%.
At the end of each epoch we anneal the softmax temperature $\tau$ to make the choice among alternative bit-widths more ``decisive''.
In all our experiments, $\tau$ is initially set to $5$ and progressively annealed by $e^{-0.0045}$ as in~\cite{wan2020fbnetv2}.
Note that in our closest prior work~\cite{cai2020rethinking}, the alternate $W$ and $\theta$ training and the softmax temperature were not present.
However, we found experimentally that both techniques improve the training stability and final result quality, not only for our approach but also for~\cite{cai2020rethinking}.

Lastly, in the \textit{fine-tuning} phase, the $\theta$ parameters are frozen to the final learned values, and the softmax is replaced with an argmax to select a single precision for each weight channel or activation layer.
Then, only the weights $W$ are fine-tuned based on $\mathcal{L_T}$.
In all our experiments, the warmup and fine-tuning epochs $\rm Epochs_{wu}$ and $\rm Epochs_{ft}$ are set equal to the training epochs reported in the papers describing the reference DNNs.
Conversely, the search phase termination is controlled with an early-stop mechanism.
\subsection{Implementation Details} \label{sec:deploy_reorg}

In this section, we show that besides having the potential to improve the \textit{theoretical} compression and efficiency of DNNs compared to standard mixed-precision, our method is also \textit{fully compatible} with existing hardware and software libraries for mixed-precision inference, with minimal overheads. 

\begin{figure}[t]
  \centering
  \includegraphics[width=1.\columnwidth]{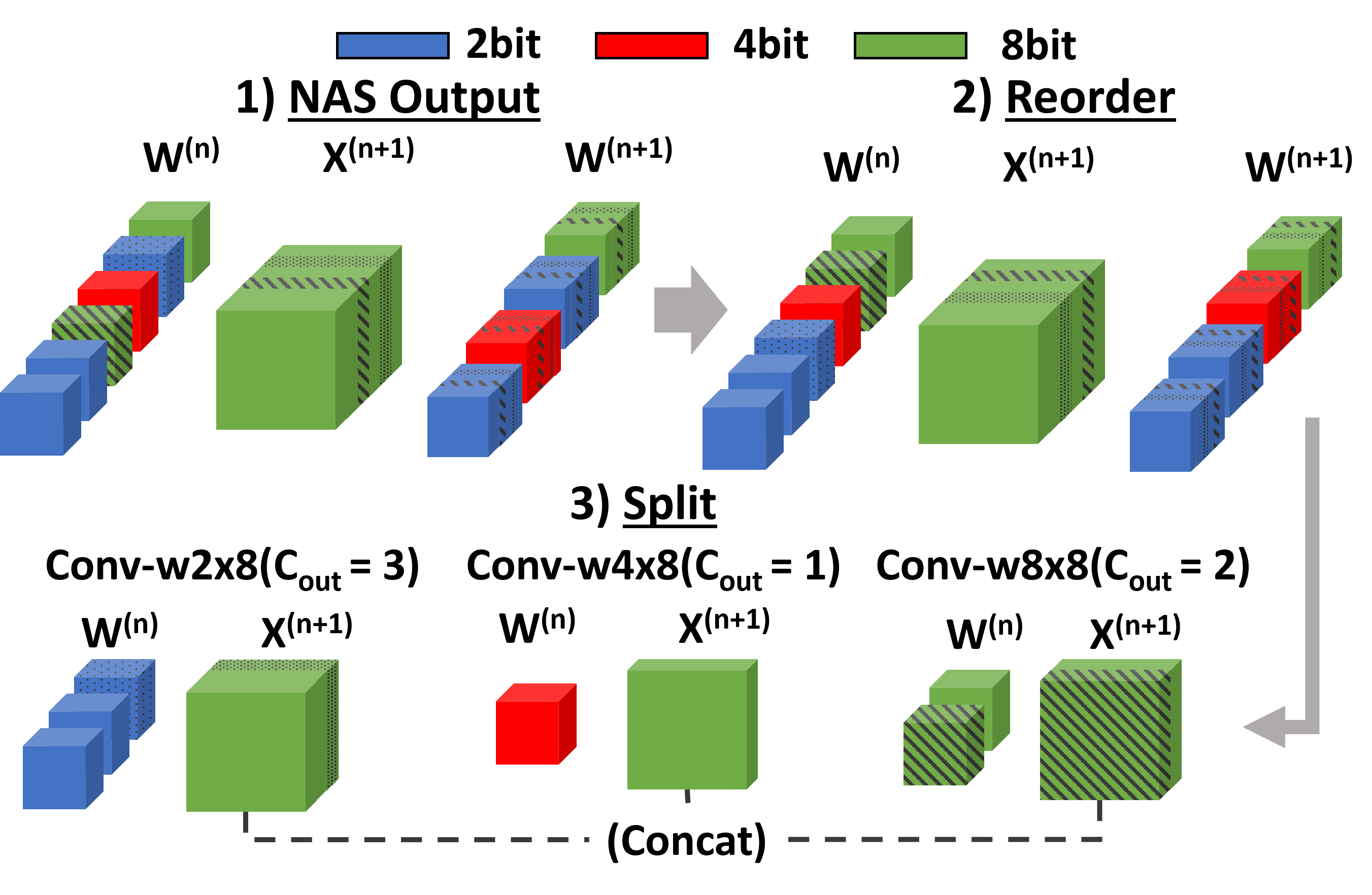}
  \vspace{-0.6cm}
  \caption{Layer re-organization to support channel-wise precision assignment.}
  \label{fig:ch-reorg}
  \vspace{-0.4cm}
\end{figure}
The top-left part of Fig.~\ref{fig:ch-reorg} shows the DNAS output for a generic Conv layer, with all filters of assigned 2,4 or 8-bit, independently from their ordering in memory (e.g., the 3rd and last filter are 8-bit, while the 4th filter is 2bit, etc). To deploy such a layer, we first \textit{reorder} the filters grouping them by bit-width (top-right of the figure). Importantly, this also affects the order of the output activations channels.  Accordingly, to preserve the functionality of the \textit{following} layer, its weight tensor has to be re-organized along the $C_{in}$ axis, so that each weight is still multiplied with the correct input activation.
In the figure, we show this associating a color pattern to the two filters that change position due to reordering, and using the same pattern to highlight the change in activations and the corresponding reorganization of the next layer weights.

After this re-organization, which is performed offline and does not have run-time overheads, the layer can be \textit{split} into $|P_W|$ separate convolutions working in parallel, each with a fraction of the original output channels. These sub-layers have a single bit-width for $W$, hence they are fully compatible with existing mixed-precision hardware and libraries~\cite{capotondi2020cmix,mpic}. Moreover, since the activation bit-width is assigned layer-wise, all outputs have the \textit{same} precision, and can be simply concatenated (i.e., stored in adjacent memory locations) to be readily usable as inputs for the next layer.
\rev{Allowing this simple concatenation is the reason why we do not perform channel-wise precision assignment for activations too. In fact, in that case the next layer's filters would have to process an interleaved mix of different precision inputs, which is not currently supported by inference libraries. Nevertheless, the study of fine-grain activation precisions will be subject of our future work. In the current version, instead,}
the only overhead of our method compared to standard mixed-precision is the control flow to schedule the three sub-layers. However, this cost is negligible compared to the benefits of the proposed fine-grain assignment.

\section{Experimental Results}\label{sec:results}
\begin{figure*}[t]
  \centering
  \includegraphics[width=.94\textwidth]{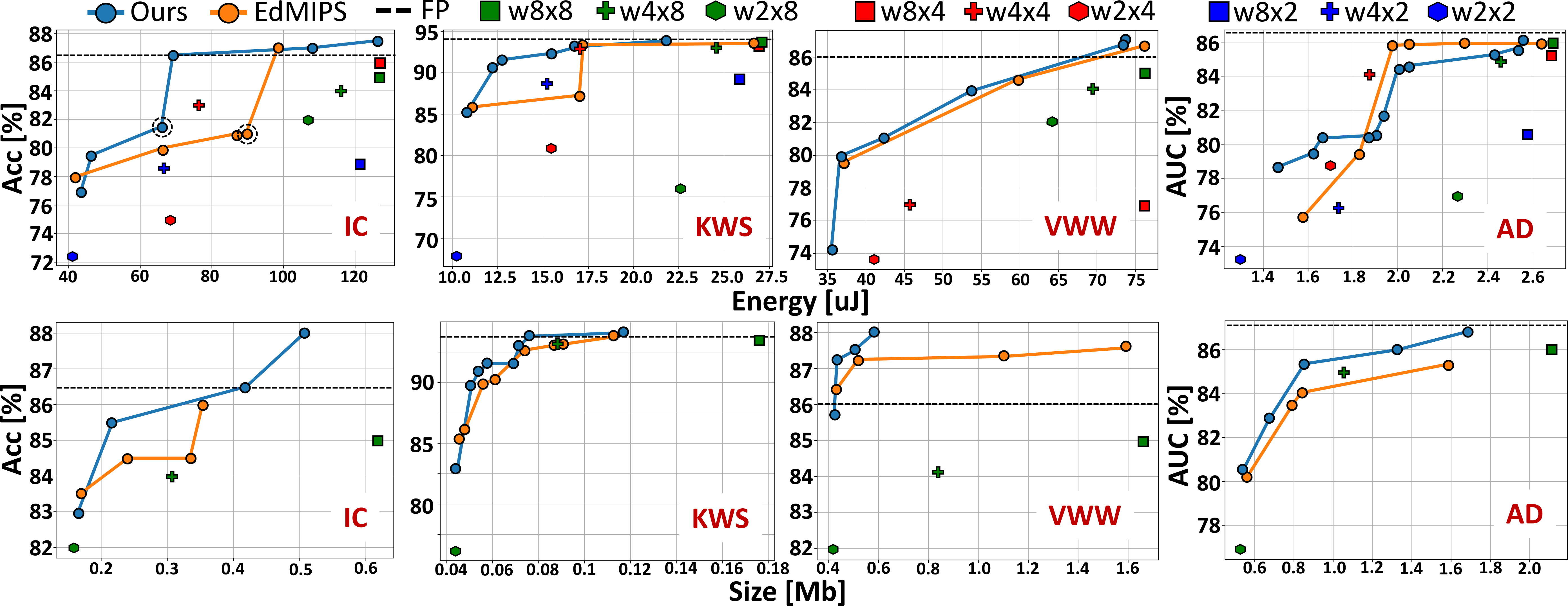}
   \vspace{-0.4cm}
  \caption{Pareto fronts obtained for the four MLPerf Tiny benchmarks, and comparison with EdMIPS and fixed-precision solutions.}
  \label{fig:pareto_fronts}
  \vspace{-0.4cm}
\end{figure*}
\subsection{Setup}
We evaluate our channel-wise mixed-precision DNAS on the four benchmarks of the MLPerf Tiny suite~\cite{banbury2021mlperf}. Below is a summary of each task, while more details can be found in~\cite{banbury2021mlperf}.
\textit{Image Classification} (IC) targets the CIFAR-10 dataset using a custom ResNet-like CNN with a backbone of 8 convolutional layers.
\textit{Keyword Spotting} (KWS) targets the Google Speech Commands (GSC) v2 dataset with a small Depthwise Separable CNN (DS-CNN) originally proposed in~\cite{zhang2017hello}.
\textit{Visual Wake Word} (VWW) uses the MSCOCO 2014 dataset for a presence detection task, solved with
a MobileNetV1~\cite{howard2017mobilenets} with a width-multiplier of 0.25.
Lastly, \textit{Anomaly Detection} (AD) targets the Toy-car subset of the DCASE2020 dataset, with a simple Dense Autoencoder.

The proposed DNAS is implemented in Python 3.9 using PyTorch v1.10.2.
Our deployment target is MPIC~\cite{mpic}, a RISC-V core including optimized hardware units for the execution of MAC operations with inputs independently quantized to $p_{w/x} \in \{2, 4, 8\}$ bit.
The LUT implementing the cost function of (\ref{eq:reg_loss_energy}) is built using the energy/OP values profiled from the MPIC core running at 250MHz, for all combinations of activations and weights precisions within the $p_{w/x}$ set.
% %

\subsection{Search-Space Exploration}
\looseness=-1
Fig.~\ref{fig:pareto_fronts} shows the results of applying our mixed-precision assignment method to the four MLPerf Tiny benchmarks. The vertical axes of all graphs report the task scores, i.e. the accuracy for IC, KWS and VWW and the Area Under the ROC Curve (AUC) for AD. The horizontal axes of the first row of graphs report the energy consumption of the models in $\mu J$, whereas those in the second row report the model sizes in Mb.

Each plot compares the mixed-precision solutions found with our tool (blue dots) with the results of EdMIPS~\cite{cai2020rethinking} (orange dots). To have a fair comparison, in which our method's advantages are solely due to the channel-wise precision assignment, we run our NAS and EdMIPS with identical training protocols, including the 20/80\% alternate $\theta$/$W$ training and the $\tau$ annealing. Moreover, we replace the original quantization algorithm of~\cite{cai2020rethinking} with \rev{PaCT~\cite{choi2018pact}, since the former would not be compatible with deployment on MPIC.}

\looseness=-1
We also compare against all relevant fixed-precision quantization baselines, denoted as ``wNxM'', where N and M are the $W$ and $X$ bit-widths taken from the $\{2, 4, 8\}$ bit set. In memory plots, we only report wNx8 baselines, since the activations bit-width is not relevant for model size. Lastly, the accuracy of a floating point version of each model is shown as a horizontal dashed line. We do not report the float model energy and size since MPIC does not have a Floating Point Unit (FPU).

Each of the Pareto-optimal points reported in the graphs for our method and for~\cite{cai2020rethinking} refers to a DNN with a different precision assignment. Multiple points are obtained changing the regularization strength $\lambda$ in (\ref{eq:dnas_loss}). We use the regularizer expressions of (\ref{eq:reg_loss_size}) and (\ref{eq:reg_loss_energy}) for memory and energy results respectively, both for our tool and for~\cite{cai2020rethinking}. \rev{Noteworthy, for all the benchmarks neither the float models nor the full 8-bit one outperform the best mixed-precision assignment, due to the well-known overfitting reduction effect of low bitwidth quantization~\cite{Jacob2018}.}

The left-most part of Fig.~\ref{fig:pareto_fronts} shows the results obtained on the IC task.
Our fine-grain mixed-precision approach outperforms all fixed-precision baselines and EdMIPS, both in terms of energy and model size.
Noteworthy, it saves up to 26.4\% energy and 35\% memory with respect to EdMIPS at iso-accuracy, while also obtaining a higher maximum accuracy, $+0.5\%$/$+1\%$ depending on the regularizer used. Similar results are obtained also for KWS (middle-left part of Fig.~\ref{fig:pareto_fronts}).
Again, our method Pareto-dominates all comparison baselines, finding solutions that save up to 27.2\% energy and 15.6\% memory for the same accuracy, and improve the best score by $+4.3\%$ and $+0.7\%$ respectively.

The middle-right part of Fig.~\ref{fig:pareto_fronts} reports the results on VWW.
On this benchmark, we Pareto-dominate fixed-precision networks both in terms of memory and energy.
Conversely, compared to EdMIPS, our method is particularly beneficial to reduce the memory footprint of the network,
with up to 63.4\% saving at equal accuracy, and a maximum accuracy improvement is $+0.4\%$.
Instead, the benefit in terms of energy is limited, although we are still able to find a \textit{larger number of Pareto-optimal points} with respect to EdMIPS, which translates to more flexibility for designers.
Note that, for this benchmark, the fixed-precision networks with 2bit activations are not shown because their training does not converge.

Lastly, the right-most part of Fig.~\ref{fig:pareto_fronts} shows the results on AD. As for all other benchmarks, our method obtains superior results in terms of AUC versus model size. The memory saving is at most 46.1\% for the same AUC with respect to EdMIPS. In this case however, \cite{cai2020rethinking} outperforms our method in terms of AUC versus energy in the high score regime (up to 21.8\% saving), while our method is superior for lower AUC values (up to 11.6\%). We attribute this result to the more difficult optimization problem solved by our tool. In fact, the AD Autoencoder is composed solely of FC layers with 128 channels (i.e., neurons), except for the bottleneck. With so many channels, the difference between the search space explored by our method and by \cite{cai2020rethinking} explodes. Hence, the gradient-based DNAS optimization likely ends up in a local minimum. Nonetheless, we think that this issue could be solved by tuning the training hyper-parameters specifically for this task, whereas in our experiments we keep all settings identical across benchmarks for fairness and reproducibility.
\subsection{Results Analysis}
\begin{figure}[t]
  \centering
  \includegraphics[width=.92\columnwidth]{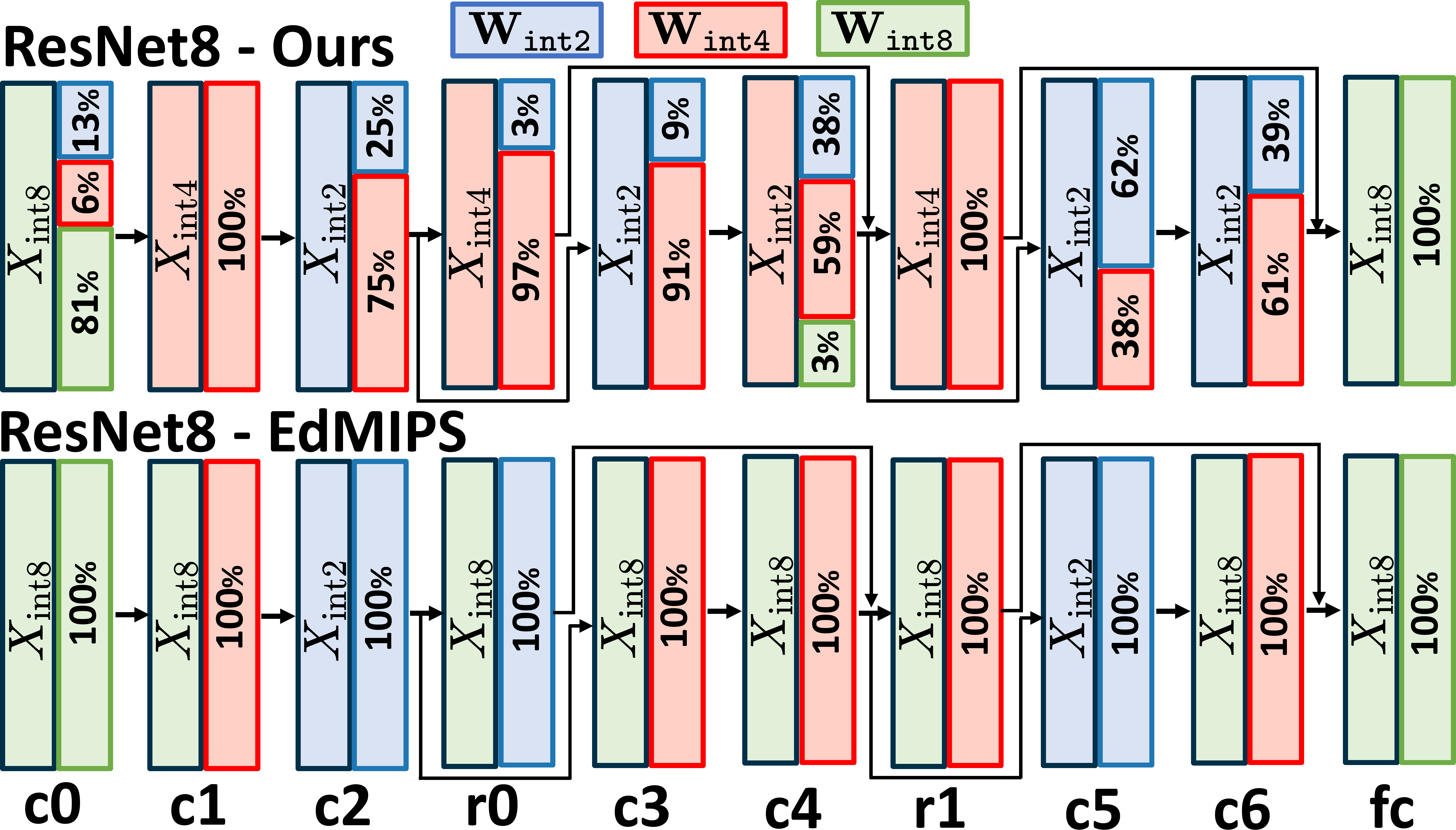}
  \vspace{-0.3cm}
  \caption{Example of found architectures for the IC benchmark.}
  \label{fig:found_arch}
  \vspace{-0.5cm}
\end{figure}
Fig.~\ref{fig:found_arch} shows an example of the precision assignments determined by our tool and by \cite{cai2020rethinking}.
The two architectures are ResNet-8 for the IC task, obtained with the energy regularizer of (\ref{eq:reg_loss_energy}), and correspond to the two circled Pareto points of Fig.~\ref{fig:pareto_fronts}, i.e., those for which our method obtains the largest energy saving with no accuracy drop.
Specifically, our channel-wise model saves 26.4\% energy with an accuracy improvement of $+0.5\%$.
Each rectangle represents a Conv (c$n$ or r$n$, where the latter are those in residual branches) or FC layer, with the activation bit-width reported on the left, and the fraction of weight channels associated to each precision on the right.

\looseness=-1
This example shows some interesting insights about the effectiveness of our approach.
For instance, it can be noticed that EdMIPS quantizes most of the activations with 8bit, whereas our method exploits its additional flexibility to \textit{reduce} the activation precision, compensating it with an increase in the bit-width assigned to an often small subset of the weights channels (e.g., only 3\% in c4), in order to obtain the same final accuracy. Eventually, only the first and last layer activations, which notoriously often require higher precision~\cite{daghero2021} remain at 8bit. Although we report a single example for sake of space, similar considerations apply to the results on the other 3 benchmarks.

As a last remark, Fig.~\ref{fig:pareto_fronts} results show that our method saves more memory than energy on MPIC. However, these memory savings could translate into further energy (and latency) reductions when considering more complex hardware, with multiple memory hierarchy levels, as larger portions of the DNN can be kept in faster and more efficient L1 memories~\cite{dory}.

\section{Conclusions}
\looseness=-1
In this work, we have proposed a new fine-grain mixed-precision search algorithm, able to efficiently select the optimal precision for each channel of each convolutional layer in CNNs.
When compared with a state-of-the-art mixed-precision DNAS, and on four edge-relevant use-cases, our tool finds richer and better trade-offs in the accuracy vs. latency/energy-consumption space, showing up to \textbf{27\%} energy consumption reduction and up to \textbf{63\%} memory reduction with no accuracy penalty.

\bibliographystyle{IEEEtran}
%\bibliography{bstctl,library}

\end{document}